\documentclass{article}

\usepackage[preprint]{neurips_2025}
\usepackage{listings}
\usepackage{xcolor}

\usepackage[utf8]{inputenc} % allow utf-8 input
\usepackage[T1]{fontenc}    % use 8-bit T1 fonts
\usepackage{hyperref}       % hyperlinks
\usepackage{url}            % simple URL typesetting
\usepackage{booktabs}       % professional-quality tables
\usepackage{amsfonts}       % blackboard math symbols
\usepackage{nicefrac}       % compact symbols for 1/2, etc.
\usepackage{amsmath}
\usepackage{microtype}      % microtypography
\usepackage{xcolor}         % colors
\usepackage{natbib}
\usepackage{wrapfig}
\usepackage{graphicx}
\usepackage[capitalize]{cleveref}

\newcommand{\myparagraph}[1]{\vspace{0pt}\noindent{\bf #1}}

\title{\emph{CraftGraffiti}: Exploring Human Identity with Custom Graffiti Art via Facial-Preserving Diffusion Models}

\author{%
  Ayan Banerjee, Fernando Vilariño, Josep Lladós \\
  Computer Vision Center, Universitat Autònoma de Barcelona\\
  \texttt{\{abanerjee, fernando, josep\}@cvc.uab.cat} \\
}

\begin{document}

\maketitle

\begin{figure}[!htbp]
%\vspace{-8mm}
\centering
\includegraphics[width=\linewidth]{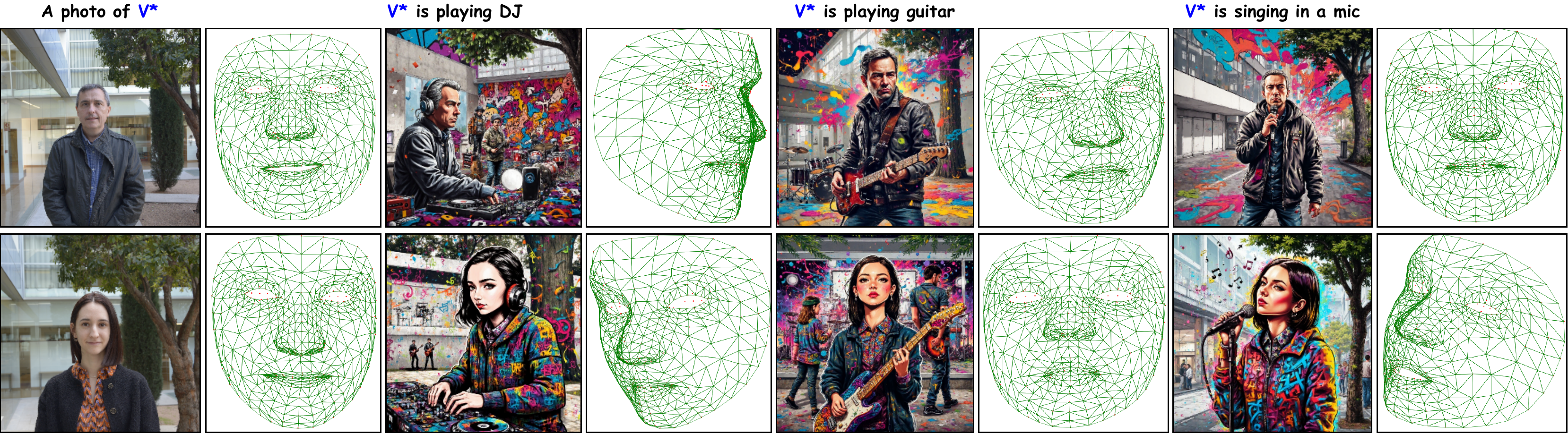}
\vspace{-20pt}
\caption{\textbf{Customization with \emph{CraftGraffiti}:} It can accurately customize the input character pose in three different musical scenarios, preserving the facial attributes in graffiti style, as the only the pose of the face mesh changes with style while preserving the mesh structure of the input image.}
\label{fig:intro}
\end{figure}

\begin{abstract}

Preserving facial identity under extreme stylistic transformation remains a major challenge in generative art. In graffiti, a high-contrast, abstract medium—subtle distortions to eyes, nose, or mouth can erase the subject’s recognizability, undermining both personal and cultural authenticity. We present \emph{CraftGraffiti}, an end-to-end text-guided graffiti generation framework designed with facial feature preservation as a primary objective. Given an input image and a style and pose descriptive prompt, \emph{CraftGraffiti} first applies graffiti style transfer via LoRA-fine-tuned pretrained diffusion transformer, then enforces identity fidelity through a face-consistent self-attention mechanism that augments attention layers with explicit identity embeddings. Pose customization is achieved without keypoints, using CLIP-guided prompt extension to enable dynamic re-posing while retaining facial coherence. We formally justify and empirically validate the “style-first, identity-after” paradigm, showing it reduces attribute drift compared to the reverse order. Quantitative results demonstrate competitive facial feature consistency and state-of-the-art aesthetic and human preference scores, while qualitative analyses and a live deployment at the Cruïlla Festival highlight the system’s real-world creative impact. \emph{CraftGraffiti} advances the goal of identity-respectful AI-assisted artistry, offering a principled approach for blending stylistic freedom with recognizability in creative AI applications. The code, demo, and details of the actual installation at the music Festival Cruilla 2025 in Barcelona are available at: \href{https://github.com/ayanban011/CraftGraffiti}{github.com/ayanban011/CraftGraffiti.}
\end{abstract}

\section{Introduction}

\myparagraph{Human identity preservation in the  GenAI era: bias, aesthetics, and open experimentation:}
Generative systems such as GANs and diffusion models replicate and amplify cultural biases from their training data, entering into the risk of distorting human identity across multiple dimensions \citep{ghosh2024don}. Since usually trained on not particularly curated datasets, it is known that current generative systems usually tend to disproportionately produce light-skinned, youthful, conventionally attractive faces, marginalizing other demographics, and concentrating on a narrow band of a specific concept of aesthetics \citep{munoz2023uncovering}. %This may particularly affect how such systems project a sexualized perspective on women and girls, due to embedded objectification patterns present both in society and in the datasets \citep{wolfe2023sexual}. 
On the other hand, both children and older adults are usually either underrepresented or not properly identified, even for age-progression models \citep{munoz2023uncovering,li2018glca}, presenting results that tend to transform the image of a young person into an adult face, or conversely,  elderly people as unrealistically younger figures. Errors in the gender identification are usually more prone to affect women than men, and reinforce gender norms and heteronormativity, as seen in occupational and presentational stereotypes \citep{sun2023smiling,zhou2024bias}. Rooted in imbalanced datasets, these distortions risk homogenizing outputs unless addressed through curated data, fairness-aware training, and interdisciplinary critique \citep{ferrara2023fairness}.

For this reason, and in light of the relevant risks identified, it is essential to openly address the tensions to which our human identity is exposed in the new digital era; an era of interconnected individuals endowed with generative capacities. Such a topic is profound and complex from the philosophical, ethical, political, and technical dimensions \citep{binns2017fairness, alkfairy2024ethical}. We state that in order to address the discussion and public debate of these topics, we can make use of two main instruments: 1) An artistic installation providing a representation of the customized human pose, technically sound and preserving facial attributes as much as possible, and 2) an ecological environment that allows for the natural interaction with people, providing the explicit context of a living lab or open experimentation space.

\myparagraph{\emph{CraftGraffiti} - Integrating Cultural Expression and Facial Consistency in Generative Street Art:}
%For the technical realization of identity-preserving generation, we choose graffiti art as our creative domain and public outreach medium. Graffiti is a culturally significant form of street art that merges vibrant expression with the diversity of human identity \citep{das2023power,forster2012evaluating}. It has historically served as a visual voice for marginalized communities \citep{bates2014bombing}, making it an ideal canvas for exploring how generative AI can inclusively represent individuals. At the same time, recent advances in generative adversarial networks (GANs) \citep{xu2021drb,yang2019controllable} and diffusion models \citep{zhang2023inversion,wang2023stylediffusion} provide powerful tools for image synthesis; yet these models still struggle to maintain detailed facial identity when confronted with drastic changes in style or pose.
For the first aim, we need a cultural bind that serves as an entry point for people to interact with this new era of GenAI, which has to be inclusive and accessible for all;  from this perspective, graffiti appears as an excellent candidate. Graffiti is a culturally significant art form, merging creative expression with the diversity of human identity \citep{das2023power,forster2012evaluating} and often serving as a voice for marginalized communities \citep{bates2014bombing}. While recent advances in GANs \citep{xu2021drb,yang2019controllable} and diffusion models \citep{zhang2023inversion,wang2023stylediffusion} offer automated generation, existing methods struggle to preserve facial attributes across varying styles and poses. Text-based image editing \citep{kawar2023imagic,huang2025diffusion} and style transfer models \citep{huang2024diffstyler,wang2023stylediffusion,guo2024focus} fail to simultaneously adjust poses, while pose editing approaches \citep{cheong2023upgpt,okuyama2024diffbody} lack unified style transfer. Unified models \citep{pascual2024enhancing,mokady2023null,dong2023prompt} attempt both but often lose facial details due to self-attention limitations.

%Achieving this combination of stylistic transformation and identity preservation remains an open technical challenge. Contemporary text-driven image editing methods \citep{kawar2023imagic,huang2025diffusion} and style transfer techniques \citep{huang2024diffstyler,wang2023stylediffusion,guo2024focus} either cannot change a subject’s pose or fail to safeguard facial details, respectively. Conversely, dedicated pose-editing frameworks \citep{cheong2023upgpt,okuyama2024diffbody} allow repositioning of a figure but cannot apply new artistic styles. Even unified generative approaches that attempt to handle both pose and style simultaneously \citep{pascual2024enhancing,mokady2023null,dong2023prompt} tend to lose fine facial features of the subject—often a consequence of self-attention mechanisms struggling to balance content (identity) and style. In practice, without special design considerations, a person’s face often becomes unrecognizable after such transformations. Addressing this issue of facial feature preservation is thus central to our work.

We introduce \emph{CraftGraffiti}, an end-to-end generative framework that produces custom graffiti-style portraits of human characters while explicitly preserving the subject’s facial identity (see \cref{fig:intro}). This system is text-guided, allowing a user to specify high-level context (e.g., prompting a portrait of the person as a DJ, guitarist, or singer) and to customize the character’s pose, all within a graffiti art style. To accomplish this, \emph{CraftGraffiti} integrates several components: a CLIP-based text encoder for semantic guidance, a graffiti-style fusion module implemented via a pre-trained diffusion transformer \footnote[1]{https://civitai.com/models/1058970/graffiti-style-fluxFaces} fine-tuned on graffiti aesthetics, a LoRA-based adapter \citep{hu2022lora} that efficiently injects pose information without requiring full model retraining, and a novel face-consistent self-attention mechanism that ensures key facial features of the input are maintained throughout the diffusion process. The final output is rendered by a variational autoencoder (VAE) decoder, yielding a high-quality graffiti portrait that retains the individual’s identity. This architecture enables simultaneous pose adjustment and style transfer while keeping the person’s face recognizable. Our contributions include: (1) a unified framework for pose-guided, graffiti-themed portrait generation that preserves individual identity; (2) a novel self-attention module specifically designed for facial consistency in diffusion-based generation; and (3) state-of-the-art performance in both qualitative visual fidelity and quantitative identity-preservation metrics compared to existing methods.

%To overcome these challenges, we present \emph{CraftGraffiti}, an end-to-end model for text-guided graffiti generation that integrates pose and style customization while preserving facial attributes (see \cref{fig:intro}). Our method employs CLIP-based encoding, a graffiti-style fusion via a pre-trained diffusion transformer\footnote[1]{https://civitai.com/models/1058970/graffiti-style-fluxFaces}, pose refinement with LoRA \citep{hu2022lora}, and a face-consistent self-attention mechanism, followed by VAE decoding. Contributions include: (1) a unified graffiti art generation framework for musical human characters; (2) a novel self-attention mechanism for facial consistency; and (3) state-of-the-art performance in quantitative and qualitative evaluations.

\myparagraph{A public artistic installation in an ecological environment as a living lab:}
For our second aim, we propose to explore ecological environments to enhance user-experience experimentation by enabling real-world engagement in authentic contexts \citep{kieffer2017ecoval}. Living labs provide user-centred, open innovation ecosystems with real-world experimentation, co-creation, and multi-stakeholder collaboration \citep{compagnucci2021livinglabs,livinglab_wikipedia}. Outdoor labs with pervasive sensing and crowdsensing further increase scalability and context-awareness \citep{pournaras2021crowdsensing}, fostering inclusivity, diversity, and richer insights than artificial environments. From this perspective, music festivals appear as excellent ecological experimentation spaces for this type of trans- and multi-disciplinary investigation. In our case, we set our installation during the four days of the Festival Cru\"illa, taking place 9-12 July 2025 (Please refer to the appendix for more details of the installation setup, outcomes, and human feedback).

\section{Preliminaries}
\myparagraph{Diffusion Models:}
Text-to-image diffusion models, such as Stable Diffusion~\cite{rombach2022high,podell2023sdxl,everaert2023diffusion}, generate images by learning a denoising process that maps latent noise to realistic images ($\mathbf{I}$), conditioned on textual input. Let $\mathbf{x}_0 \in \mathbb{R}^{H \times W \times C}$ denote a clean image and $\mathbf{x}_t$ its noisy version at timestep $t \in \{1, \dots, T\}$. The forward diffusion process ($q (\cdot)$) adds Gaussian noise ($\mathcal{N} (\cdot)$) in a Markov chain:
\begin{equation}
    q(\mathbf{x}_t \mid \mathbf{x}_{t-1}) = \mathcal{N}(\mathbf{x}_t; \sqrt{1 - \beta_t} \mathbf{x}_{t-1}, \beta_t \mathbf{I}),
\end{equation}
with a predefined variance schedule \( \{\beta_t\} \). The model learns the conditional reverse process ($p_\theta$):
\begin{equation}
    p_\theta(\mathbf{x}_{t-1} \mid \mathbf{x}_t, \mathbf{c}) = \mathcal{N}(\mathbf{x}_{t-1}; \boldsymbol{\mu}_\theta(\mathbf{x}_t, t, \mathbf{c}), \boldsymbol{\Sigma}_\theta(\mathbf{x}_t, t, \mathbf{c})),
\end{equation}
where $\mathbf{c}$ is a conditioning vector, often a CLIP-encoded text prompt. Training minimizes the error between predicted and true noise or the original image $\mathbf{x}_0$, depending on parameterization (mean: $\boldsymbol{\mu}_\theta$, and covariance $\boldsymbol{\Sigma}_\theta$). Latent diffusion models (LDMs) \citep{wu2023latent,nam2024contrastive,zhang2023sine} perform this process in a compressed latent space \( \mathcal{Z} \), using an encoder \( \mathcal{E}: \mathbb{R}^{H \times W \times C} \rightarrow \mathcal{Z} \) and decoder \( \mathcal{D}: \mathcal{Z} \rightarrow \mathbb{R}^{H \times W \times C} \), improving computational efficiency while preserving generation quality.

\begin{figure}[!htbp]
%\vspace{-8mm}
\centering
\includegraphics[width=\textwidth]{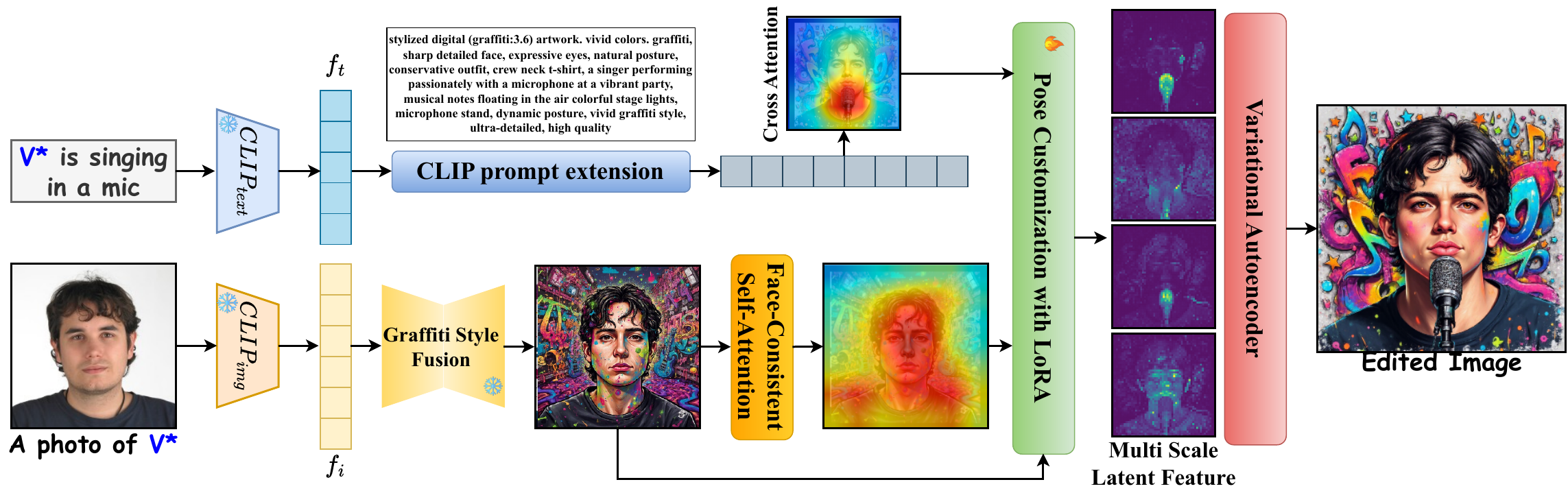}
%\vspace{-8mm}
\vspace{-20pt}
\caption{\textbf{\emph{CraftGraffiti}} transforms a source image into a graffiti-style portrait while preserving the subject’s identity and pose. Graffiti style is injected via a pretrained diffusion fine-tuned with LoRA for the dedicated style. Later on, another diffusion model is equipped with face-consistent self-attention and cross-attention modules to preserve key facial features, and a LoRA module enables pose customization without full model retraining via CLIP-based prompt extension. Finally, multi-scale latent feature processing using a VAE ensures that both global structure and fine details are captured across different resolutions in the latent space, yielding a high-quality graffiti-style image.}
\label{fig:method}
%\vspace{-4mm}
\end{figure}

\myparagraph{Style Transfer:}
Style transfer \citep{wang2023interactive,hamazaspyan2023diffusion,everaert2023diffusion} aims to modify the appearance of a generated image $\mathbf{x}$ to match the style of a reference image $\mathbf{s}$, while maintaining the content from a source image $\mathbf{c}$. In feature space $\phi(\cdot)$, the style and content losses are defined as:
\begin{equation}
    \mathcal{L}_{\text{style}}(\mathbf{x}, \mathbf{s}) = \sum_{l} \left\| G^{(l)}(\phi^{(l)}(\mathbf{x})) - G^{(l)}(\phi^{(l)}(\mathbf{s})) \right\|_F^2,
\end{equation}
\begin{equation}
    \mathcal{L}_{\text{content}}(\mathbf{x}, \mathbf{c}) = \left\| \phi^{(l)}(\mathbf{x}) - \phi^{(l)}(\mathbf{c}) \right\|_2^2,
\end{equation}
where $G^{(l)}$ denotes the Gram matrix \cite{li2017universal} of features at layer $l$, capturing second-order statistics. The total loss combines both objectives:
\begin{equation}
    \mathcal{L}_{\text{total}} = \lambda_c \mathcal{L}_{\text{content}} + \lambda_s \mathcal{L}_{\text{style}},
\end{equation}
with tunable weights $\lambda_c$ and $\lambda_s$. Modern approaches integrate style conditioning into generative models via cross-attention modulation or adapter modules. We fine-tune the transformer of the FLUX.1 dev \footnote[2]{https://huggingface.co/black-forest-labs/FLUX.1-dev} with low-rank adaptation via the $\mathcal{L}_{\text{total}}$ in order to learn the graffiti style and use it as a pre-trained model during inference to pose the style over the input image (see \cref{fig:method}).

\myparagraph{Pose Customization with LoRA:}
Generation of characterization conditions on a specified human pose $\mathbf{p}$, often represented as a 2D keypoint skeleton or heatmap \citep{wang2024stable,ju2023humansd,khandelwal2024reposedm}. This signal can be embedded and injected into a diffusion model via attention layers or early-stage concatenation. LoRA \citep{hu2022lora} offers an efficient means of adapting large diffusion models to new tasks (e.g., pose-guided generation) by injecting low-rank updates into pretrained weight matrices. For a weight matrix $W \in \mathbb{R}^{d \times k}$, LoRA introduces trainable low-rank matrices $A \in \mathbb{R}^{d \times r}$, $B \in \mathbb{R}^{r \times k}$, modifying the weights as:
\begin{equation}
    \tilde{W} = W + \alpha \cdot AB,
\end{equation}
where $\alpha$ is a scaling factor and $r \ll \min(d, k)$. By freezing base weights and only training $A$ and $B$, LoRA allows efficient fine-tuning with significantly fewer parameters. LoRA can be applied to attention matrices $W_q, W_k, W_v$, enabling fast pose-driven customization without full model retraining. In \emph{CraftGraffiti} instead of using keypoints or heatmap, we extend the input prompt via T5 \citep{narang2021transformer} and get the cross-attention map as a guided signal ($B$) for the pose (see \cref{fig:method}). We maintain the identity of the character through the face-consistent self-attention ($A$) and update the self and cross attention of the diffusion transformer via fine-tuning with LoRA for multi-scale latent feature generation. This multi-scale latent feature is further denoised via VAE in order to get the final edited image in graffiti style with an accurate musical pose described in the text prompt.

\section{CraftGraffiti}
\label{sec:03}
\subsection{Why should we perform graffiti style transfer before facial attribute preservation}
Performing the style transform before enforcing attribute constraints can be seen via a simple model. Let $I$ be the original face image and let $S(I)$ be its graffiti version under a LoRA-fine-tuned diffusion model. If we have a facial-attribute extractor $F_a(I)\in\mathbb{R}^k$ (encoding eyes, nose, chin, expression, etc.) and an attribute-loss $\mathcal{L}_{attr} (X) = ||F_a(X) - F_a(I)||^2$, we can choose a projection $P(\cdot)$ so that for any image $X$, $F_a(P(X))=F_a(I)$ (i.e. $P$ restores $I$’s attributes). Then by reconstruction $L_{\rm attr}(P(S(I)))=0$, whereas $P(I)=I$ (since $I$ already has its own attributes), so $\mathcal{L} (S(P(I))) = \mathcal{L}(S(I)) = ||F_a(S(I)) - F_a(I)||^2 > 0$ whenever $S$ perturbs the attributes. In other words, $\mathcal{L}_{attr} (P \circ S(I)) \le \mathcal{L}_{attr} (S \circ P(I))$. This shows formally that applying the style $S$ first and then the attribute correction $P$ yields lower attribute discrepancy. This matches intuition from prior work: style‐transfer methods aim to alter appearance while preserving its underlying structure \citep{rezaei2025training}; however, if the style influence is too strong, it undermines the structural integrity of the content \citep{wang2024instantstyle}. Moreover, LoRA fine-tuned diffusion models often use the early timesteps to reconstruct coarse content and later steps to add stylistic detail \citep{ouyang2025k}. Thus, we preserve identity better by first adding the graffiti style and finally preserving the facial features, rather than applying constraints before styling. The overall methodology of \emph{CraftGraffiti} has been demonstrated in \cref{fig:method}.

\textbf{Theorem.} Let $I$ be the input face with attributes $a=F_a(I)$. Let $S$ be a diffusion‐style operator and $P$ an operator satisfying $F_a(P(X))=a$ for all $X$ (so $P$ restores $I$’s attributes). Then the image $X'=P(S(I))$ obeys $F_a(X')=a$ (it exactly preserves the original attributes), whereas $X''=S(P(I))$ satisfies $F_a(X'')=F_a(S(I))$ and in general $F_a(S(I))\neq a$ unless $S$ itself preserved $F_a$. Hence $(P\circ S)(I)$ maintains the facial attributes exactly, whereas $(S\circ P)(I)$ need not.

\textbf{Proof.} By assumption $F_a(P(X))=a=F_a(I)$ for any input $X$. In particular, $F_a(P(S(I)))=a$, so $P(S(I))$ has the original attribute vector. On the other hand, since $P(I)=I$ (the identity image already has attributes $a$), we have $S(P(I))=S(I)$ and hence $F_a(S(P(I)))=F_a(S(I))$. Unless $S$ is attribute-preserving by itself, $F_a(S(I))\neq F_a(I)$ in general. Therefore, $P(S(I))$ preserves $a$ exactly, while $S(P(I))$ typically does not, completing the proof.

\subsection{How the extra dimension in the self-attention helps in facial attribute preservation}
\begin{wrapfigure}{r}{0.5\textwidth}
%\vspace{-4mm}
\centering
\includegraphics[width=0.5\textwidth]{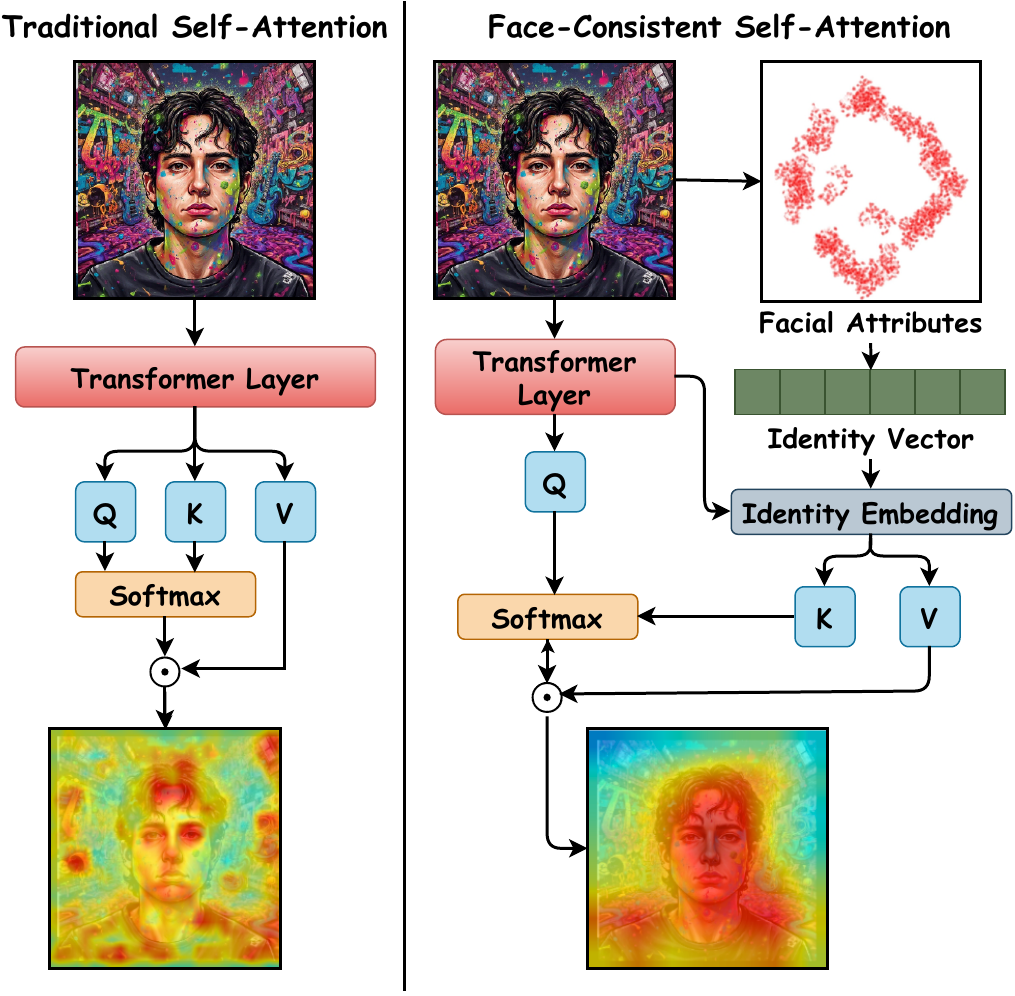}
%\vspace{-8mm}
\caption{\textbf{face-consistent self-attention:} we can easily preserve the facial attribute of the character through the extra dimension of identity embedding.}
\label{fig:self}
\vspace{-4mm}
\end{wrapfigure}

In Diffusion models \citep{zhou2024storydiffusion,wang2024autostory}, each self-attention layer computes attention only among features of the current latent (no cross-latent computation). Concretely, the latent feature map of size $H \times W$ is reshaped into a sequence of tokens, and for each token, a query ($Q$), key ($K$), and value ($V$) vector are computed from the same latent features. The attention weights are then obtained by taking dot products of $Q$ and $K$ across all positions (scaled by affinity distance $\sqrt{d}$) and passing through softmax, and the result reweights the $V$ vectors (see \cref{fig:self}). This means every pixel in the latent try to see every other pixel in the same latent when computing self-attention.

To ensure that the two generated characters depict the same identity, we introduce an explicit identity embedding into the attention computation. Conceptually, this means adding a special identity vector (often derived from a reference face) as an extra token or feature dimension in the self-attention layer. In practice, one can concatenate an identity embedding to every spatial token (equivalently, add an extra “CLS”-style token) or append an identity channel to the latent features. The result is that queries and keys now carry identity information as well as image content. In effect, each pixel’s query includes a fixed identity code, and each key (or value) can be augmented by the same code, biasing attention towards features matching that identity.

In an identity-enhanced self-attention, the attention matrix is computed using both the spatial features and the identity embedding. For example, one might form $Q' = W_Q[x; id]$, $K' = W_K[x; id]$ where x is a spatial feature and id is the identity vector. Then, attention weights come from $Q'K'^T$. This extra dimension (the identity channel or token) causes the softmax to favor aligning features that correspond to the same identity. Architecturally, the U-Net is unchanged except at attention layers: we simply extend the input feature dimension by the identity embedding. The key effect is that the same identity code is shared across the entire image and across images in a batch. As a result, when generating multiple images of the same person, the self-attention layers see the same identity embedding each time, forcing consistent facial features.

\section{Experimental Results}
\myparagraph{Implementation Details:}
We implement \emph{CraftGraffiti} in PyTorch \cite{paszke2019pytorch}, utilizing a pre-trained .1 dev \footnote[3]{https://huggingface.co/spaces/black-forest-labs/FLUX.1-dev} (12 billion parameter) transformer specifically designed for graffiti style transfer. The guidance scale is set to  7.5. For preserving facial attributes while fine-tuning the transformer with low rank adaptation, the subject guidance factor ($\lambda$) is set to 0.95, and the style intensity factor is set to 0.7. We set the rank ($\sigma$) to 64 and the regularization factor ($\alpha$) to 128 while fine-tuning the transformer for pose customization. We run the denoising process with 100 steps by default. We only perform latent composition in the first quarter of the denoising process (the first 25 steps). All experiments ran on a single NVIDIA A6000 48GB GPU.

\myparagraph{Dataset and Evaluation Metric:}
For graffiti style generation, we have used the images of the 17K-Graffiti dataset \footnote[4]{https://github.com/visual-ds/17K-Graffiti} to fine-tune our style fusion diffusion with LoRA and use it as a pre-trained model during inference. Also, to further validate \emph{CraftGraffiti}, we have gathered images of people from our laboratory, the Computer Vision Center, Barcelona \footnote[5]{https://www.cvc.uab.es/}, with their consent, and use them in the paper for validation and pose customization.
Furthermore, to perform a fair qualitative analysis, we test \emph{CraftGraffiti} and other state-of-the-art approaches with facial feature consistency (FFC) (implementation in the Appendix), aesthetic score (Aes) \citep{murray2012ava} calculated using LAION aesthetic classifier, and Human Preference Score (HPS) \citep{wu2023human}. Also, we have compared them against the inference time to understand their real-time use case scenario.

\begin{figure}[!htbp]
%\vspace{-8mm}
\centering
\includegraphics[width=\textwidth]{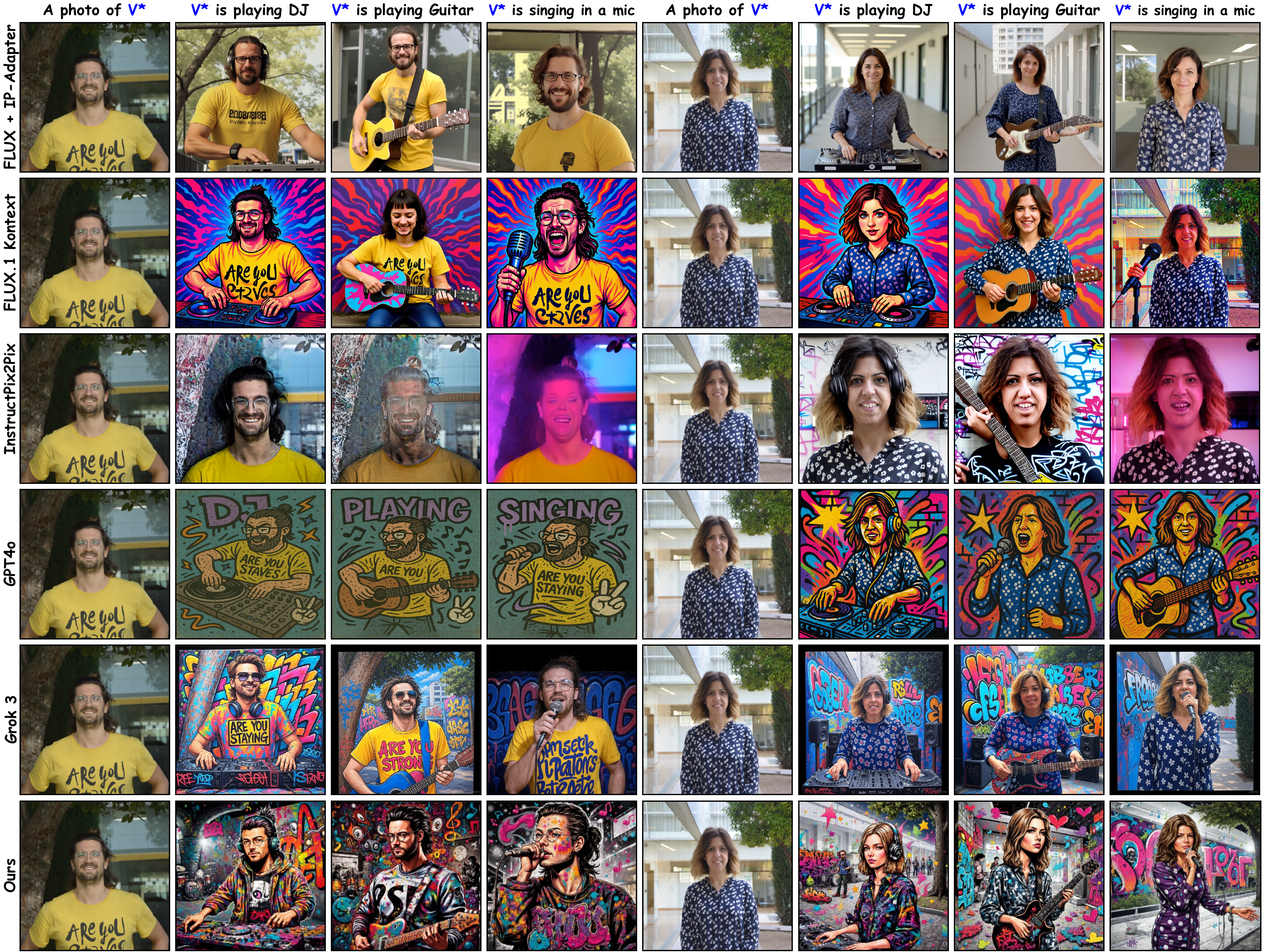}
%\vspace{-20pt}
\caption{\textbf{Qualitative Comparison:} CraftGraffiti perfectly transforms the input image into the graffiti style and maintains facial attributes, while the rest cannot do both.}
\label{fig:qual}
\vspace{-4mm}
\end{figure}

\myparagraph{Qualitative Analysis:}
We compare \emph{CraftGraffiti} against several FLUX-based approaches, FLUX + IP-Adapter \citep{ye2023ip}, and FLUX.1 Kontext \citep{labs2025flux}, however, both approaches fail to generate the graffiti style. Although FLUX + IP-Adapter \citep{ye2023ip} tried to maintain facial consistency, it was unable to change style. On the other hand, FLUX.1 Kontext \citep{labs2025flux} tries to change the background into a more cartoonist style, and the faces are not consistent. This shows how hard to unify the style transfer and consistency tasks and justify our decision to perform style transfer before facial consistency (see \cref{fig:qual}).

We also compare \emph{CraftGraffiti} with InstructPix2Pix \citep{brooks2023instructpix2pix}; however, it neither adds objects nor generates high-quality images. Similarly, VLMs (Grok 3 \citep{grok3beta} and GPT4o \citep{hurst2024gpt}) maintain consistency, unable to blend style due to the complexity of graffiti style transfer. This empirically shows the complexity of the problem (facial consistency in graffiti style) we are tackling and demonstrates the correctness of the component choice of \emph{CraftGraffitii} for accurate style blending by preserving the facial features (For more qualitative examples, please refer to the \cref{fig:qualm} of the appendix.)

\myparagraph{Quantitative Analysis:}
\cref{tab:01} presents a quantitative comparison against the state-of-the-art image editing techniques. \emph{CraftGraffiti} stands in the fourth position after GPT4 \citep{hurst2024gpt}, GPT4o \citep{hurst2024gpt}, and \citep{ye2023ip} respectively, in terms of facial feature consistency (FFC), as it is very difficult to compare the preservation of facial features, as most of the feature extraction models (InceptionResNetV1) consider global features rather than the local ones. However, during style transfer, it is hard to preserve the global features of the input images. As other image editing \citep{grok3beta,hurst2024gpt,ye2023ip} models are unable to blend the graffiti style, preserve the global features, and perform better in this metric.

\begin{wraptable}{r}{0.5\textwidth}
\vspace{-5mm}
\centering
\caption{Quantitative Evaluation of CraftGraffiti}
\label{tab:01}
\resizebox{0.5\textwidth}{!}{
\begin{tabular}{@{}lcccc@{}}
\toprule
Method/Metric         & FFC $\uparrow$    & Aes $\uparrow$   & HPS $\uparrow$   & Inf. Time (sec) $\downarrow$\\ \midrule
FLUX + IP-Adapter \citep{ye2023ip}    & 0.8324 & 3.2414 & 0.3012 & 8.2             \\
FLUX.1 Kontext \citep{labs2025flux}        & 0.6741 & 2.1749 & 0.2911 & 6.7             \\
InstructPix2Pix \citep{brooks2023instructpix2pix}      & 0.7112 & 2.7272 & 0.1918 & 3.1             \\
GPT4o \citep{hurst2024gpt}                & 0.8761 & 4.5193 & 0.3412 & 13.4            \\
Grok 3 \citep{grok3beta}               & 0.8513 & 4.1652 & 0.3102 & 11.2            \\ \midrule
Ours (Baseline)       & 0.7618 & 3.6913 & 0.2911 & 2.9             \\
Ours (+ Style Fusion) & 0.6814 & 4.7195 & 0.3001 & 5.3             \\
\begin{tabular}[c]{@{}c@{}}Ours (+ Face Consistent \\ self attention)\end{tabular} & 0.7713 & 5.1376 & 0.3176 & 8.7 \\
CraftGraffiti         & 0.7713 & 5.2271 & 0.3536 & 10.1            \\ \bottomrule
\end{tabular}}
\vspace{-4mm}
\end{wraptable}

On the other hand, Aesthetic Score \citep{murray2012ava,banerjee2025svgcraft} and Human Preference Score \citep{wu2023human} measure how perfectly the models blend the graffiti style with the input images. \emph{CraftGraffiti} outperforms all the state-of-the-art image editing techniques in both metrics, as they are unable to blend the graffiti style properly over the input images. Last but not least, \emph{CraftGraffiti} has very little inference time compared to the VLMs \citep{grok3beta,hurst2024gpt}, which makes it suitable for real-time applications. %Further ablation studies of \emph{CraftGraffiti} are available in the \cref{fig:ab} and \cref{fig:abs} of the appendix.

\begin{wrapfigure}{r}{0.65\textwidth}
\vspace{-8mm}
\centering
\includegraphics[width=0.65\textwidth]{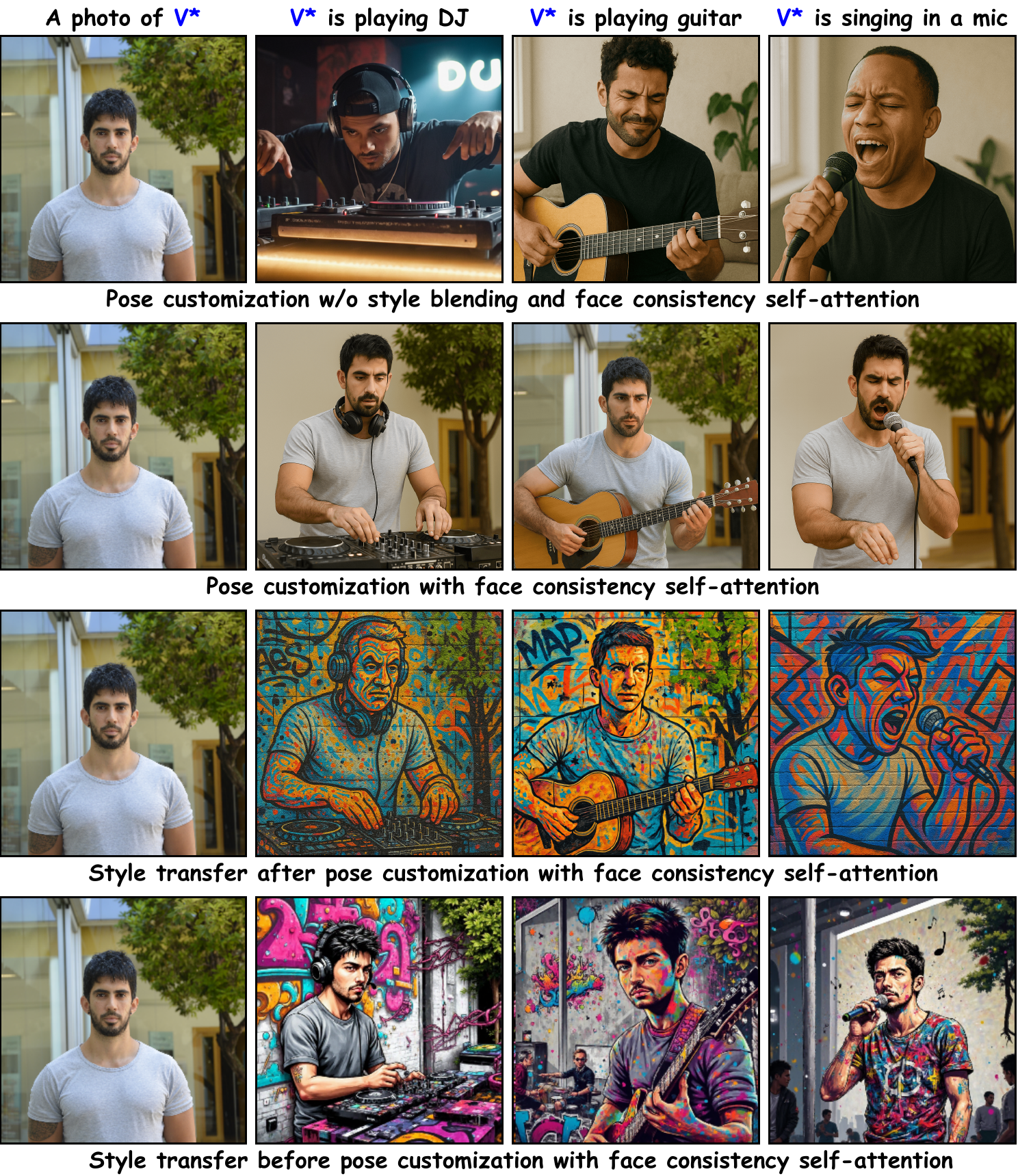}
%\vspace{-20pt}
\caption{\textbf{Ablation Studies:} With face consistency self-attention, we preserve facial attributes during pose customization, and with primary style blending, the graffiti posters appear more realistic.}
\label{fig:ab}
\vspace{-4mm}
\end{wrapfigure}

\myparagraph{Ablation Studies:}
In order to understand the importance of style blending before the pose customization and the face-consistency self-attention, we have performed a brief ablation study depicted in \cref{fig:ab}. It has been observed that with the FLUX.1 dev (12B) baseline, we neither achieve consistency nor achieve the style transfer. On the other hand, we have applied the face consistency self-attention before applying the style transfer. It has been observed that it can preserve the facial attribute during pose customization with low-rank adaptation (2nd row of \cref{fig:ab}). Similarly, if we perform style transfer after the pose customization, neither the facial attributes are preserved, even with the presence of face-consistent self-attention, nor does the style reflect graffiti. As depicted in the 3rd row of \cref{fig:ab}, the images look like a cartoonish drawing on a wall rather than a graffiti poster. So the correct paradigm is to perform the style blending first and then pose customization with face consistent self-attention (4th row of \cref{fig:ab}).

\begin{wrapfigure}{r}{0.60\textwidth}
\vspace{-4mm}
\centering
\includegraphics[width=0.60\textwidth]{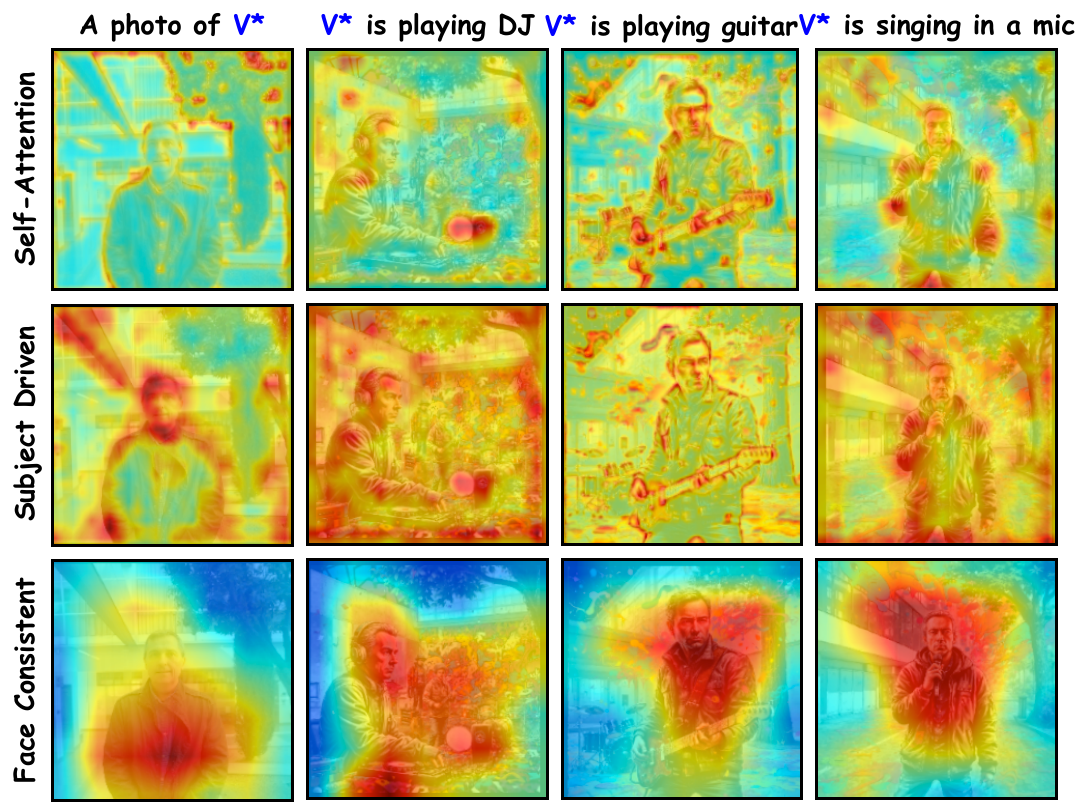}
%\vspace{-8mm}
\caption{\textbf{Ablation of the self-attention:} The face consistent self-attention primarily focuses on the human faces and their corresponding poses, whereas the traditional self-attention and subject-driven self-attention of Consistory \citep{tewel2024training} diverges towards the global scenario.}
\label{fig:abs}
\vspace{-4mm}
\end{wrapfigure}

Similarly, we perform an ablation study on the self-attention (see \cref{fig:abs}) to preserve the facial attribute during pose customization to understand why the extension of the attention dimension helps to preserve the facial features. In order to do that, we took the generated images with \emph{CraftGraffiti} and pass them through the self-attention layer of FLUX.1 dev, Subject-driven self-attention layer of Consistory \citep{tewel2024training}, and our face consistent self-attention layer. It has been observed that the traditional self-attention of FLUX.1 dev focuses on a certain point, whereas Subject-driven Self-Attention of ConsiStory \citep{tewel2024training} attends the global features rather than the local ones. On the other hand, our proposed face-consistent self-attention, focused on the local features of the facial attributes, helps to maintain facial attributes during face customization.
 
\begin{wrapfigure}{r}{0.68\textwidth}
\vspace{-4mm}
\centering
\includegraphics[width=0.68\textwidth]{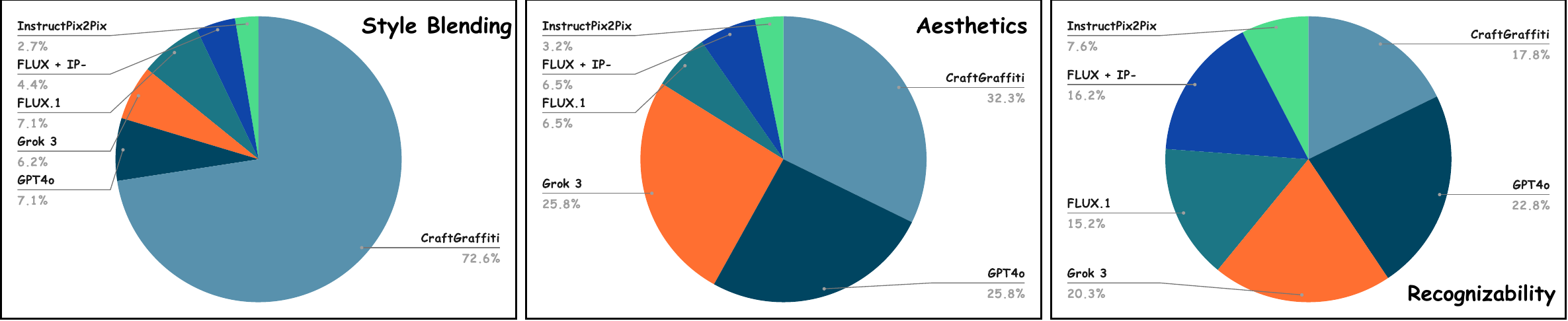}
%\vspace{-8mm}
\caption{\textbf{Human Evaluation:} \emph{CraftGraffiti} outperforms SOTA techniques \citep{ye2023ip,labs2025flux,brooks2023instructpix2pix,grok3beta,hurst2024gpt} in style blending and aesthetics while preserving facial attributes (a decent performance in recognizability).}
\label{fig:human}
\vspace{-3mm}
\end{wrapfigure}

\myparagraph{Human Evaluation:}
We have conducted the perceptual study in order to evaluate the credibility and widespread adoption of the graffiti poster synthesized with \emph{CraftGraffiti}. We chose 60 graffiti posters (20 for each pose) synthesized by \citep{ye2023ip,labs2025flux,brooks2023instructpix2pix,grok3beta,hurst2024gpt} and \emph{CraftGraffiti} and asked the users to rate the graffiti on a scale of 1 to 5 (1 Low; 5 high) based on aesthetics, style blending, and recognizability. 47 anonymous users participated in this study. The outcome of this user study is reported in \cref{fig:human}.

For the user, \emph{CraftGraffiti} makes a perfect graffiti-style poster, while other models either change the background or make it more cartoonish. In terms of recognizability \emph{CraftGraffiti} preserves the local features (eyes, nose, lips, and so on) but changes the global features during style transfer, maintaining a decent recognizability compared to the rest.

\section{Discussion on \emph{CraftGraffiti} Performance}

\myparagraph{Technical Contributions and Performance:}
Our results demonstrate that integrating a face-consistent self-attention mechanism with LoRA-based fine-tuning allows \emph{CraftGraffiti} to preserve key facial attributes across diverse poses and graffiti styles. Compared to existing pose-editing and style-transfer methods, \emph{CraftGraffiti} achieves state-of-the-art identity preservation without compromising artistic quality. This balance between recognizability and stylization is essential for applications where personal identity and cultural authenticity are intertwined.

\myparagraph{Cultural and Societal Implications:}
Beyond technical performance, \emph{CraftGraffiti} acts as an enabler in ongoing cultural debates around representation, bias, and access to AI-generated art. By using graffiti, a medium historically tied to marginalized voices, we highlight the potential for generative systems to serve as inclusive cultural tools. However, these systems also risk reproducing and amplifying harmful biases, which must be addressed through dataset curation, fairness-aware training, and critical engagement with affected communities. By deploying the installation at Festival Cru\"illa, we had an opportunity to observe user interactions in a high-energy, socially diverse, real-world context. This aligns with the principles of ecological validity \citep{kieffer2017ecoval} and living lab methodologies \citep{compagnucci2021livinglabs,livinglab_wikipedia}. The festival setting facilitated spontaneous engagement, enabling us to capture authentic emotional responses and uncover interaction patterns that may not emerge in controlled lab settings \citep{pournaras2021crowdsensing}. Such environments also pose technical challenges in terms of managing environmental noise, variable lighting, unpredictable crowd dynamics, and regulatory uncertainties. In our specific case, it appeared as an excellent example to dry test the new European AI Act. The implications of this regulatory learning approach are of enormous interest, though they are beyond the scope of this piece of work.

\myparagraph{Limitations:}
While our model successfully preserves facial attributes, %it remains constrained by the biases present in its training data, which may affect demographic and stylistic diversity. Furthermore, 
real-time generation in public settings is computationally demanding, and user interactions are influenced by the physical constraints of the installation space. Our evaluation was conducted in a single cultural context (a European music festival), which may limit the generalizability of the findings to other socio-cultural settings. However, general considerations can be taken regarding the questions posed: \emph{CraftGraffiti} produces graffiti-styled human figures that preserve the original subject’s identity by maintaining key facial attributes across poses and styles through a face-consistent self-attention mechanism. Unlike many generative systems, it inherits body shape from the input image, avoiding homogenization into a single aesthetic norm. From an anthropological perspective, the tendency of many AI models to generate light-skinned, symmetrical, youthful, and slender bodies reflects the reinforcement of dominant, often Eurocentric, beauty standards \citep{munoz2023uncovering}. Such homogenization risks erasing diversity in age, body size, ability, and ethnic features, mirroring historical processes where global media displaced local aesthetics. %Additionally, the sexualization of women’s bodies \citep{wolfe2023sexual} and gendered facial expressions \citep{sun2023smiling} embed cultural power asymmetries into outputs. 
Addressing these issues requires combining computational audits with ethnographic insights to prevent generative AI from fostering an aesthetic monoculture. %A relevant example of this is the actual sexualization of the image of younger people, particularly in the case of women, were the generative model can project an image corresponding to the potentially future body shape, attached to sexual maturity which may or may not be present, acting as an intriguing time machine and a graphical reflection tool for the sexual dimension. 
Last but not least, it is of crucial importance to highlight the need for a strict control on the access of children, since, even though the interaction with the installation could be perceived as an entertainment action both from the children's and parent's perspective, it could create an actual unexpected impact in a moment of body development with psychological uncertainties. 
%The CraftGraffiti system generates graffiti-style human figures that preserve subjects’ facial identity and body shape, avoiding the homogenization common in many generative models. However, it inherits biases from training data, limiting demographic and stylistic diversity, and faces computational and spatial constraints for real-time public use. The evaluation, conducted in a single European cultural context, may not generalize globally. Anthropologically, generative AI often reinforces Eurocentric beauty standards, erases diversity, sexualizes women, and embeds gendered expressions, perpetuating cultural power imbalances. A specific risk is the sexualization of images of younger individuals, where the system could project adult-like features onto children, potentially causing psychological harm. Strict access controls are essential to protect minors, as seemingly playful interactions may have unintended developmental impacts.

\myparagraph{Future Directions:}
Future work is definitly needed, and potential lines or work will have to explore: (1) extending \emph{CraftGraffiti} to handle a broader range of cultural art forms beyond graffiti; (2) integrating real-time bias detection and mitigation pipelines; (3) testing in varied ecological environments, including community art events and educational settings; and (4) expanding cross-cultural studies to assess how identity preservation and stylistic adaptation are perceived in different cultural contexts.

\section{Conclusion}
\emph{CraftGraffiti} represents a significant step forward in the intersection of generative AI and urban art, enabling the creation of personalized graffiti that preserves facial identity while embracing stylistic transformation. By introducing a face-consistent self-attention mechanism and leveraging LoRA-based fine-tuning, the system ensures that key facial features such as the eyes, nose, and mouth remain consistent across pose and style changes, addressing the longstanding challenge of identity preservation in diffusion-based image generation. This technical innovation allows for high-quality graffiti renderings that are both aesthetically compelling and faithful to the original subject, blending artistic stylization with personal recognizability. Beyond its technical contributions, \emph{CraftGraffiti} holds broader societal and artistic implications by democratizing access to street art and empowering individuals and communities to participate in shaping their visual environments. By maintaining cultural motifs and individual likenesses, the model enables the creation of meaningful, identity-rich urban expressions that reflect diverse human experiences. As a result, \emph{CraftGraffiti} not only advances the state of the art in generative customization but also offers a transformative tool for preserving and celebrating cultural narratives through the accessible medium of AI-generated graffiti.

%\section{Acknowledgements}
%This piece of research was carried out with the support of the following grant projects:
%SGR Grant 2021 SGR 01559 from the Catalan Government, GRAIL PID2021-126808OB-I00, and SUKIDI PID2024-157778OB-I00 grants from the Spanish Ministry of Science and Innovation, and with the support of Cátedra UAB-Cruïlla grant TSI-100929-2023-2 from the Ministry of Economic Affairs and Digital Transformation of the Spanish Government. 

%\section*{References}
%\newpage
\bibliographystyle{plain}
\bibliography{references}

\newpage
%%%%%%%%%%%%%%%%%%%%%%%%%%%%%%%%%%%%%%%%%%%%%%%%%%%%%%%%%%%%

\appendix

\section{Demonstration at Cruïlla Festival}
\begin{figure}[!htbp]
%\vspace{-8mm}
\centering
\includegraphics[width=\textwidth]{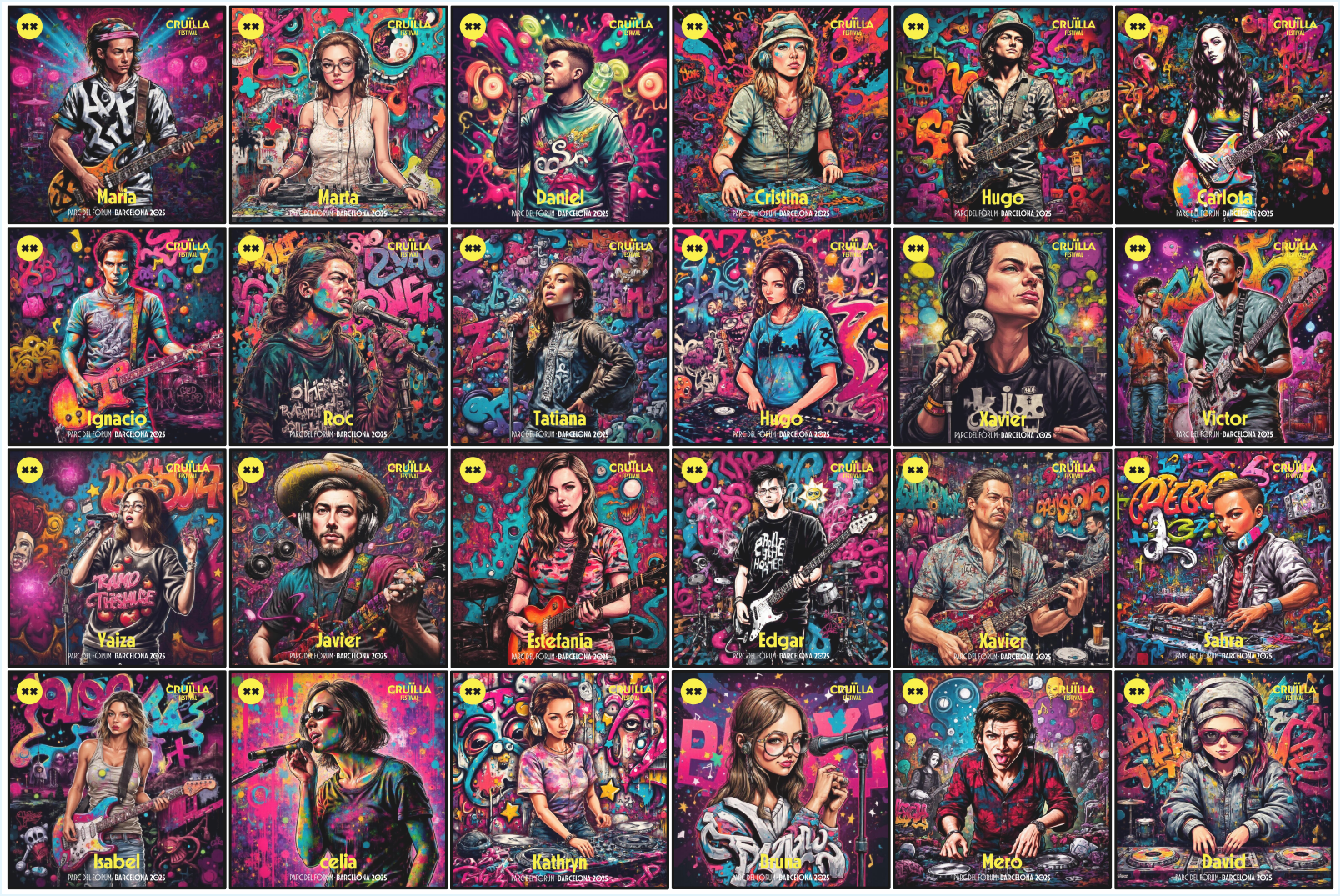}
%\vspace{-8mm}
\caption{Example outcomes from the demonstration at Cruilla Festival Barcelona 2025}
\label{fig:cruilla}
%\vspace{-8mm}
\end{figure}

An installation (see \cref{fig:inst}) implementing the proposed system was deployed during the Festival Cruïlla (www.cruillabarcelona.com) in Barcelona from 9-12 July 2025. The festival hosts 80,000 people during 4 days, attending parallel contexts and activities related to culture and innovation. More than 1,100 people visited the booth in which the installation was set, and 586 people were able to create their impression of a graffiti-style poster (see \cref{fig:cruilla}) and give their feedback anonymously. We have summarized the human feedback and listed it as follows:

\begin{figure}[!htbp]
%\vspace{-10mm}
\centering
\includegraphics[width=\textwidth]{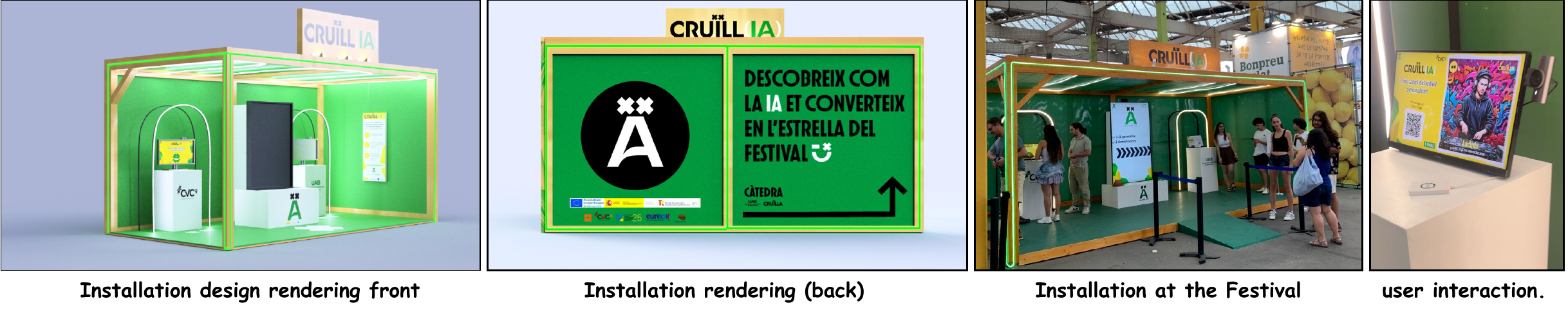}
%\vspace{-8mm}
\caption{\textbf{\emph{CraftGrafitti} artsitic installation:} Conceptual rendering and Actual setup at Cru\"illa Festival 2025}
\label{fig:inst}
%\vspace{-8mm}
\end{figure}

(1) The general impression of the users was that the demo was fun and engaging, often reacting with surprise and amusement at the results. The majority of participants understood that the demonstration was intended as a playful, exploratory experience rather than a precise or professional tool.

(2) Some users asked if their photos were being stored or used to train the algorithm, highlighting the importance of transparency in data management. We informed that none of the user's data has been stored, referring to the informed consent regarding the use of the images only for quality control. The output images of \cref{fig:cruilla} are are obtained with the permission of the user to showcase the output of the generation for solely research purpose.

(3)The system tended to apply heavy makeup to women’s faces, regardless of whether the user was wearing any in the original photo. Generated images of women occasionally included exaggerated features, such as full lips and prominent cheekbones.

(4) A large number of participants commented that the results looked quite similar and their identity was reflected by the system. This was also consistent when asked about their opinion on how the system represented other participants, across users, with only slight variations in hairstyle or clothing, reducing the perceived uniqueness.

(5) The system tended to make people look younger, except for the younger people, in which an explicit evolution on the perceived maturity was identified, particularly in aspects related to body shape, such as muscular development for men or stylized forms for women.

(6) Even though gender consistency is high, for a limited number of cases, some women subjects were interpreted as men. In this case, the user was immediately addressed by the installation operators to validate the reaction and re-state the experimental nature of the installation, addressing the particular misfunctions that it can have and providing valuable feedback for the improvement.

\section{Some more qualitative examples}
\begin{figure}[!htbp]
%\vspace{-16mm}
\centering
\includegraphics[width=\textwidth]{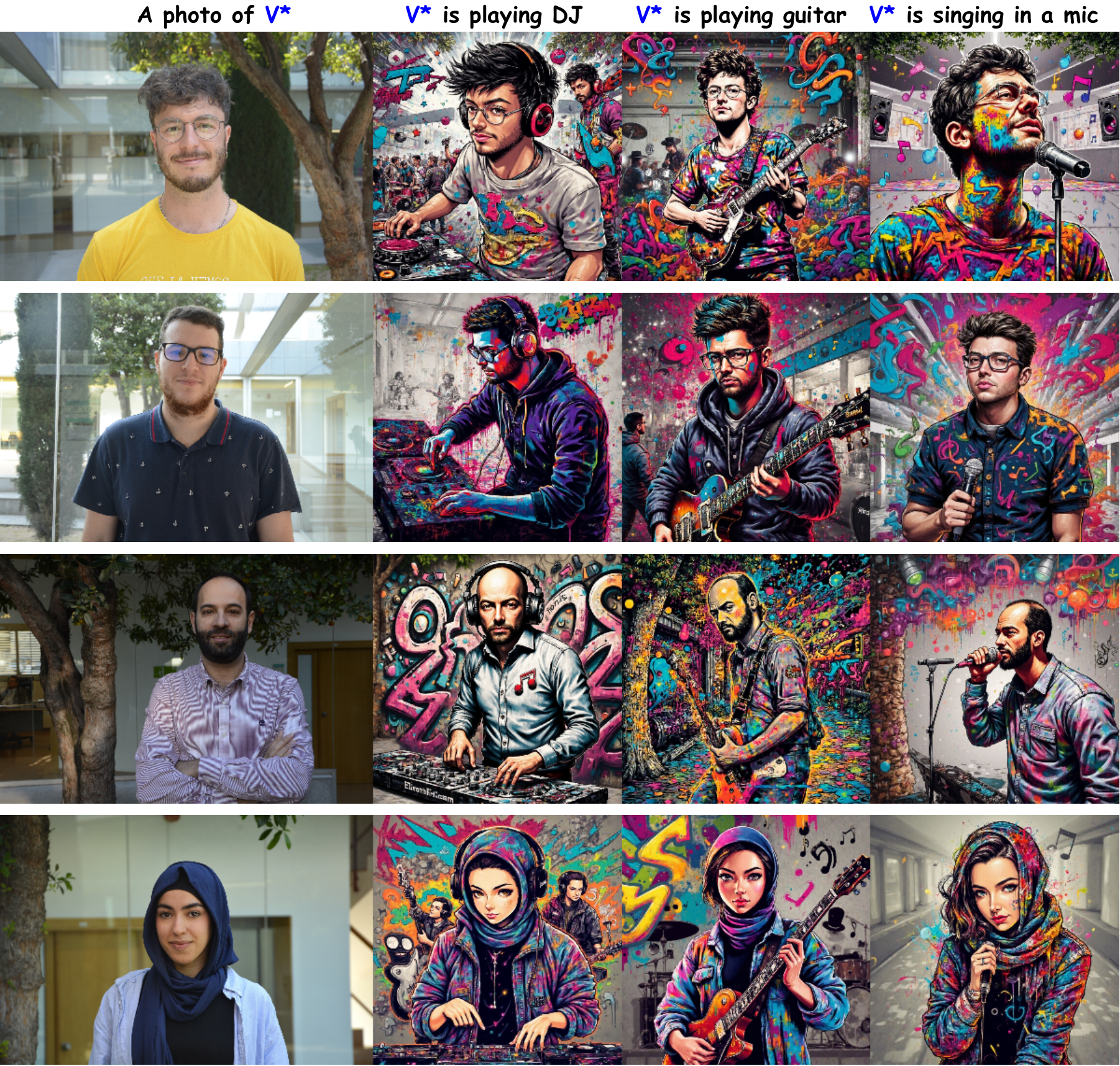}
%\vspace{-8mm}
\caption{Some more qualitative examples generated with \emph{CraftGraffiti} maintaining facial attribute with proper style blending.}
\label{fig:qualm}
>%\vspace{-16mm}
\end{figure}

We generated some more qualitative examples with \emph{CraftGraffiti} as reported in \cref{fig:qualm}. It has been observed that, if the user uses some external facial accessories (e.g., classes) \emph{CraftGraffiti} also preserves it in graffiti style (1st and 2nd row of \cref{fig:qualm}). Similarly, if the user is bald or wearing hijab, it maintains that hair pattern and the external clothing accessories too (3rd and 4th row of \cref{fig:qualm}). 
 
\section{Future Directions}
Future work is definitly needed, and potential lines or work will have to explore: (1) extending \emph{CraftGraffiti} to handle a broader range of cultural art forms beyond graffiti; (2) integrating real-time bias detection and mitigation pipelines; (3) testing in varied ecological environments, including community art events and educational settings; and (4) expanding cross-cultural studies to assess how identity preservation and stylistic adaptation are perceived in different cultural contexts.

\section{Ethics Statement}
The application of \emph{CraftGraffiti} in image editing offers extensive potential for diverse downstream applications, facilitating the adaptation of images to different contexts. The primary objective of our model is to automate and streamline this process, resulting in significant time and resource savings. It is important to acknowledge that current methods have inherent limitations, as discussed in this paper. However, our model can serve as an intermediary solution, expediting the creation process and offering valuable insights for further advancements. It is crucial to remain mindful of potential risks associated with these models, including the dissemination of misinformation, potential for abuse, and introduction of biases. Broader impacts and ethical considerations should be thoroughly addressed and studied in order to responsibly harness the capabilities of such models. Terms and conditions forms were explicitly offered to all the participants during the live festival, and researchers and support team were always present during the realization of the images, vigilant to potential issues appearing and aware of the user's reactions.

\section{Code for facial consistency evaluation}
Through the following implementation, we first load the FaceNet \citep{schroff2015facenet} model pretrained on VGGFace2 \citep{cao2018vggface2}, which is designed to produce numerical vector embeddings for faces. Given two embedding vectors
\[
\mathbf{u}, \mathbf{v} \in \mathbb{R}^n
\]
the cosine similarity is defined as:
\[
\text{cosine\_similarity}(\mathbf{u}, \mathbf{v}) =
\frac{\mathbf{u} \cdot \mathbf{v}}
{\|\mathbf{u}\|_2 \, \|\mathbf{v}\|_2}
\]
where:
\begin{itemize}
    \item $\mathbf{u} \cdot \mathbf{v}$ is the dot product of the two vectors.
    \item $\|\mathbf{u}\|_2$ and $\|\mathbf{v}\|_2$ are their Euclidean norms.
\end{itemize}

A value close to \( 1 \) indicates that the two embeddings are highly similar in direction, meaning the corresponding faces are likely to be of the same identity. A value near \( 0 \) indicates no similarity, while negative values imply opposite feature orientations.
\newpage
\begin{lstlisting}[language=Python, caption={Computing cosine similarity between two face embeddings}]
from PIL import Image
import numpy as np
import torch
from facenet_pytorch import InceptionResnetV1
import torchvision.transforms as T
from sklearn.metrics.pairwise import cosine_similarity
import matplotlib.pyplot as plt

# Load FaceNet model
model = InceptionResnetV1(pretrained='vggface2').eval()

# Load and preprocess images
transform = T.Compose([
    T.Resize((160, 160)),
    T.ToTensor(),
    T.Normalize([0.5], [0.5])
])

img1 = transform(Image.open("face1.png")).unsqueeze(0)
img2 = transform(Image.open("face2.png")).unsqueeze(0)

# Get embeddings
emb1 = model(img1).detach().numpy()
emb2 = model(img2).detach().numpy()

# Cosine similarity
similarity = cosine_similarity(emb1, emb2)[0][0]
print(f"Cosine similarity: {similarity:.4f}")
\end{lstlisting}

%%%%%%%%%%%%%%%%%%%%%%%%%%%%%%%%%%%%%%%%%%%%%%%%%%%%%%%%%%%%
%\clearpage
%\newpage
%\input{tex/08_checklist}

\end{document}